\documentclass{article}
\usepackage{ijcai18}

\usepackage{times}
\usepackage{xcolor}
\usepackage{soul}
\usepackage[utf8]{inputenc}
\usepackage[small]{caption}
\usepackage{graphicx}
\usepackage{amsmath}
\usepackage{amssymb}
\usepackage{amsthm}
\usepackage{multirow, booktabs}
\usepackage{makecell}
\usepackage{relsize}
\usepackage{hyperref}
\usepackage{subcaption}
\newcommand{\citet}[1]{\citeauthor{#1}~\shortcite{#1}}

\newcommand{\iflong}[1]{\textbf{}}

\usepackage{array}
\usepackage{color}
\newcolumntype{L}[1]{>{\raggedright\let\newline\\\arraybackslash\hspace{0pt}}m{#1}}

\title{Aspect Term Extraction with History Attention and Selective Transformation\thanks{The work was done when Xin Li was an intern at Tencent AI Lab. The project is substantially supported by a grant from the Research Grant Council of the Hong Kong Special Administrative Region, China (Project Code: 14203414).}}

\author{
Xin Li$^1$, 
Lidong Bing$^2$, 
Piji Li$^1$,
Wai Lam$^1$,
Zhimou Yang$^3$
\\ 
$^1$Key Laboratory of High Confidence Software Technologies, Ministry of Education (CUHK Sub-Lab), \\ Department of Systems Engineering and Engineering Management, \\The Chinese University of Hong Kong, Hong Kong \\
$^2$Tencent AI Lab, Shenzhen, China\\
$^3$College of Information Science and Engineering, Northeastern University, China\\
\{lixin, wlam, pjli\}@se.cuhk.edu.hk,
lyndonbing@tencent.com, yangzhimou@stumail.neu.edu.cn \\
}

\begin{document}

\maketitle

\begin{abstract}
Aspect Term Extraction (ATE), a key sub-task in Aspect-Based Sentiment Analysis, aims to extract explicit aspect expressions from online user reviews. We present a new framework for tackling ATE. It can exploit two useful clues, namely opinion summary and aspect detection history. Opinion summary is distilled from the whole input sentence, conditioned on each current token for aspect prediction, and thus the tailor-made summary can help aspect prediction on this token. Another clue is the information of aspect detection history, and it is distilled from the previous aspect predictions so as to leverage the coordinate structure and tagging schema constraints to upgrade the aspect prediction. Experimental results over four benchmark datasets clearly demonstrate that our framework can outperform all state-of-the-art methods.\footnote{Codes and datasets are available at \url{https://github.com/lixin4ever/HAST}.} 
\end{abstract}

\section{Introduction}
Aspect-Based Sentiment Analysis (ABSA) involves detecting opinion targets and locating opinion indicators in sentences in product review texts~\cite{liu2012sentiment}. The first sub-task, called 
Aspect Term Extraction (ATE), is to identify the phrases targeted by opinion indicators in review sentences. For example, in the sentence ``\textit{I love the operating system and preloaded software}'', the words ``operating system'' and ``preloaded software'' should be extracted as aspect terms, and the sentiment on them is conveyed by the opinion word ``love''. According to the task definition, for a term/phrase being regarded as an aspect, it should co-occur with some ``opinion words'' that indicate a sentiment polarity on it~\cite{pontiki-EtAl:2014:SemEval}.


Many researchers formulated ATE as a sequence labeling problem or a token-level classification problem. Traditional sequence models such as Conditional Random Fields (CRFs)~\cite{chernyshevich:2014:SemEval,toh-wang:2014:SemEval,toh-su:2016:SemEval,yin2016unsupervised}, Long Short-Term Memory Networks (LSTMs)~\cite{liu-joty-meng:2015:EMNLP} and classification models such as Support Vector Machine (SVM)~\cite{manek2016aspect} have been applied to tackle the ATE task, and achieved reasonable performance. One drawback of these existing works is that they do not exploit the fact that, according to the task definition, aspect terms should co-occur with opinion-indicating words. Thus, the above methods tend to output false positives on those frequently used aspect terms in non-opinionated sentences, e.g., the word ``restaurant'' in ``\textit{the restaurant was packed at first, so we waited for 20 minutes}'', which should not be extracted because the sentence does not convey any opinion on it. 


There are a few works that consider opinion terms when tackling the ATE task. \cite{wang-EtAl:2016:EMNLP20164} proposed Recursive Neural Conditional Random Fields (RNCRF) to explicitly extract aspects and opinions in a single framework. Aspect-opinion relation is modeled via joint extraction and dependency-based representation learning. One assumption of RNCRF is that dependency parsing will capture the relation between aspect terms and opinion words in the same sentence so that the joint extraction can benefit. Such assumption is usually valid for simple sentences, but rather fragile for some complicated structures, such as clauses and parenthesis. Moreover, RNCRF suffers from errors of dependency parsing because its network construction hinges on the dependency tree of inputs. 
CMLA~\cite{wang2017coupled} models aspect-opinion relation without using syntactic information. Instead, it enables the two tasks to share information via attention mechanism. For example, it exploits the global opinion information by directly computing the association score between the aspect prototype and individual opinion hidden representations and then performing weighted aggregation. However, such aggregation may introduce noise. To some extent, this drawback is inherited from the attention mechanism, as also observed in machine translation~\cite{luong-pham-manning:2015:EMNLP} and image captioning~\cite{xu2015show}.



To make better use of opinion information to assist aspect term extraction, we distill the opinion information of the whole input sentence into opinion summary\footnote{Technically, opinion summary is the linear combination of the opinion representations generated from LSTM.}, and such distillation is conditioned on a particular current token for aspect prediction. Then, the opinion summary is employed as part of features for the current aspect prediction. Taking the sentence ``\textit{the restaurant is cute but not upscale}'' as an example, when our model performs the prediction for the word ``restaurant'', it first generates an opinion summary of the entire sentence conditioned on ``restaurant''. Due to the strong correlation between ``restaurant' and ``upscale'' (an opinion word), the opinion summary will convey more information of ``upscale'' so that it will help predict ``restaurant'' as an aspect with high probability.
Note that the opinion summary is built on the initial opinion features coming from an auxiliary opinion detection task, and such initial features already distinguish opinion words to some extent. Moreover, we propose a novel transformation network that helps strengthen the favorable correlations, e.g. between ``restaurant' and ``upscale'', so that the produced opinion summary involves less noise.

Besides the opinion summary, another useful clue we explore is the aspect prediction history due to the inspiration of two observations: (1) In sequential labeling, the predictions at the previous time steps are useful clues for reducing the error space of the current prediction. For example, in the B-I-O tagging (refer to Section \ref{sec:task}), if the previous prediction is ``O'', then the current prediction cannot be ``I''; (2) It is observed that some sentences contain multiple aspect terms. For example, \textit{``Apple is unmatched in product quality, aesthetics, craftmanship, and customer service''} has a coordinate structure of aspects. Under this structure, the previously predicted commonly-used aspect terms (e.g., ``product quality'') can guide the model to find the infrequent aspect terms (e.g., ``craftmanship''). To capture the above clues, our model distills the information of the previous aspect detection for making a better prediction on the current state.

Concretely, we propose a framework for more accurate aspect term extraction by exploiting the opinion summary and the aspect detection history. Firstly, we employ two standard Long-Short Term Memory Networks (LSTMs) for building the initial aspect and opinion representations recording the sequential information. To encode the historical information into the initial aspect representations at each time step, we propose truncated history attention to distill useful features from the most recent aspect predictions and generate the history-aware aspect representations. We also design a selective transformation network to obtain the opinion summary at each time step. Specifically, we apply the aspect information to transform the initial opinion representations and apply attention over the transformed representations to generate the opinion summary. Experimental results show that our framework can outperform state-of-the-art methods. 

\begin{figure*}[!t]
\centering
\includegraphics[width=0.8\textwidth]{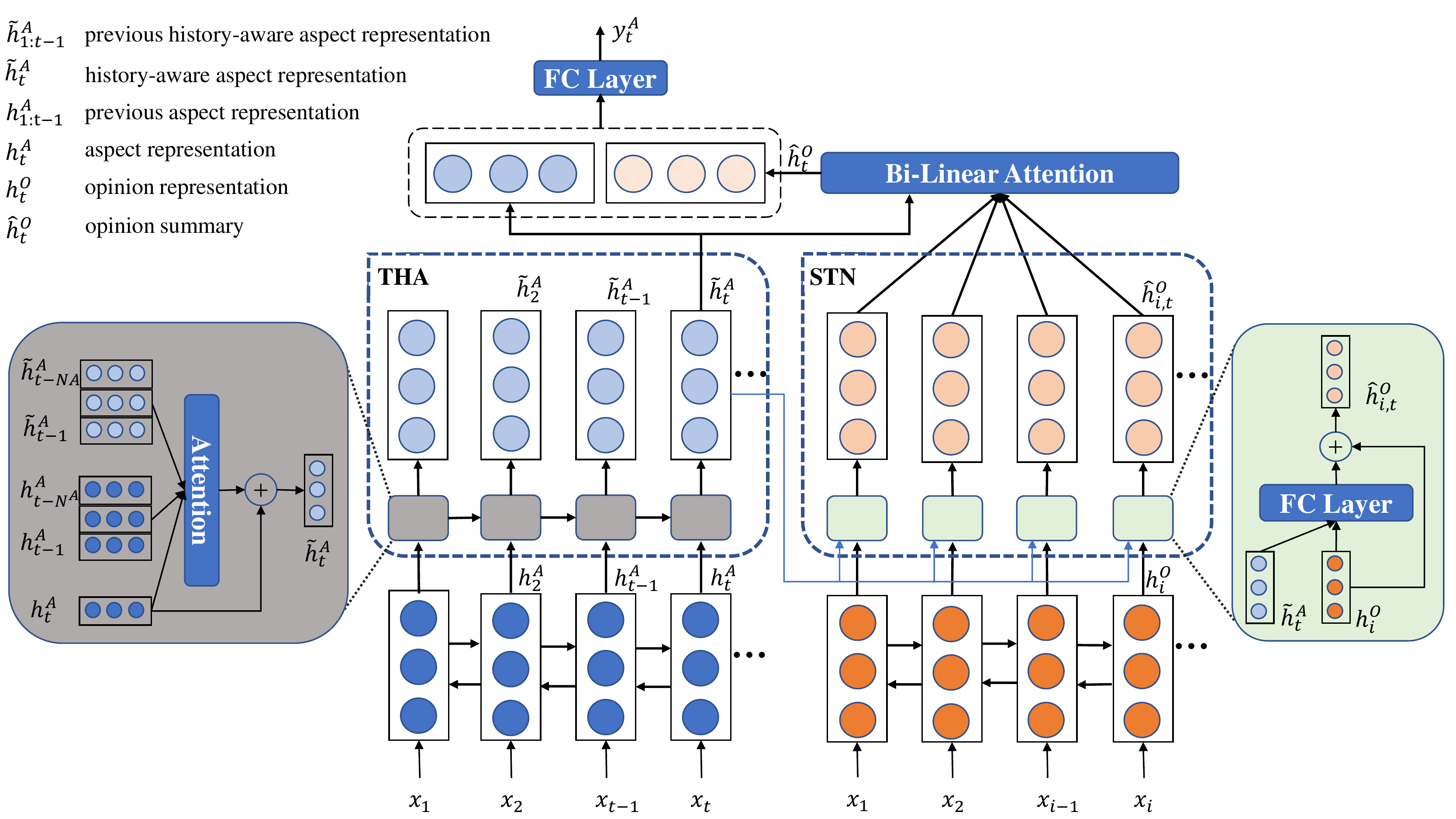}
\caption{Framework architecture. The callouts on both sides describe how THA and STN work at each time step. Color printing is preferred.}
\label{fig:architecture}
\end{figure*}

\section{The Proposed Model}
\subsection{The ATE Task}
\label{sec:task}  
Given a sequence $X = \{x_1,...,x_{T}\}$ of $T$ words, the ATE task can be formulated as a token/word level sequence labeling problem to predict an aspect label sequence $Y = \{y_1,...,y_{T}\}$, where each $y_i$ comes from a finite label set $\mathcal{Y}=  \{B, I, O\}$ which describes the possible aspect labels. As shown in the example below:\\
\resizebox{0.48\textwidth}{!}{
\begin{tabular}{c|cccccccc}
\hline
$X$ & I & love & the & operation & system & and & preloaded & software \\ \hline
$Y$ & O & O & O & B & I & O & B & I  \\ \hline
\end{tabular}}
\noindent $B$, $I$, and $O$ denote beginning of, inside and outside of the aspect span respectively. Note that in commonly-used datasets such as \cite{pontiki-EtAl:2016:SemEval}, the gold standard opinions are usually not annotated.


\subsection{Model Description}
As shown in Figure \ref{fig:architecture}, our model contains two key components, namely \textbf{T}runcated \textbf{H}istory-\textbf{A}ttention (THA) and  \textbf{S}elective \textbf{T}ransformation \textbf{N}etwork (STN), for capturing aspect detection history and opinion summary respectively. 
THA and STN are built on two LSTMs that generate the initial word representations for the primary ATE task and the auxiliary opinion detection task respectively. 
THA is designed to integrate the information of aspect detection history into the current aspect feature to generate a new history-aware aspect representation. STN first calculates a new opinion representation conditioned on the current aspect candidate. Then, we employ a bi-linear attention network to calculate the opinion summary as the weighted sum of the new opinion representations, according to their associations with the current aspect representation. 
Finally, the history-aware aspect representation and the opinion summary are concatenated as features for aspect prediction of the current time step.

\subsubsection{Building Memory}
As Recurrent Neural Networks can record the sequential information~\cite{graves2012supervised}, we employ two vanilla LSTMs to build the initial token-level contextualized representations for sequence labeling of the ATE task and the auxiliary opinion word detection task respectively. 
\iflong{The computation process of both LSTMs is identical to that in~\cite{jozefowicz2015empirical}.
\begin{equation}
\label{eq:lstm}
\begin{split}
\begin{bmatrix}
i_t \\
f_t \\
\hat{c}_t \\
o_t
\end{bmatrix}
&=
\begin{bmatrix}
\sigma \\
\sigma \\
\text{tanh} \\
\sigma
\end{bmatrix}
\begin{bmatrix}
\mathrm{\bf W} & \mathrm{\bf U} & I
\end{bmatrix}
\begin{bmatrix}
x_t \\
\tilde{h}_t\\
\mathrm{\bf b}
\end{bmatrix}, \\
c_t &= i_t \odot \hat{c}_t + c_{t-1} \odot f_t, \\
h_t &= \text{tanh}(c_t) \odot o_t,
\end{split}
\end{equation}
where $\mathrm{\bf W}$, $\mathrm{\bf U}$ and $\mathrm{\bf b}$ are learnable parameters.}
For simplicity, let $\text{LSTM}^{\mathcal{T}}(x_t)$ denote an LSTM unit where $\mathcal{T} \in \{A, O\}$ is the task indicator. In the following sections, without specification, the symbols with superscript $A$ and $O$ are the notations used in the ATE task and the opinion detection task respectively. We use Bi-Directional LSTM to generate the initial token-level representations $h^{\mathcal{T}}_t \in \mathbb{R}^{2\mathrm{dim}^{\mathcal{T}}_h}$ ($\mathrm{dim}^{\mathcal{T}}_h$ is the dimension of hidden states):
\begin{small}
\begin{equation}
h^{\mathcal{T}}_t = [\overrightarrow{\text{LSTM}}^{\mathcal{T}}(x_t); \overleftarrow{\text{LSTM}}^{\mathcal{T}}(x_t)], t \in [1, T].
\end{equation}
\end{small}
\subsubsection{Capturing Aspect History}
In principle, RNN can memorize the entire history of the predictions~\cite{graves2012supervised}, but there is no mechanism to exploit the relation between previous predictions and the current prediction. As discussed above, such relation could be useful because of two reasons: (1) reducing the model's error space in predicting the current label by considering the definition of B-I-O schema, (2) improving the prediction accuracy for multiple aspects in one coordinate structure. 

We propose a Truncated History-Attention (THA) component (the \textbf{THA} block in Figure~\ref{fig:architecture}) to explicitly model the aspect-aspect relation. Specifically, THA caches the most recent $N^{A}$ hidden states. At the current prediction time step $t$, THA calculates the normalized importance score $s^t_i$ of each cached state $h^A_i$ ($i \in [t-N^A, t-1]$) as follows:
\begin{small}
\begin{equation}
    a^t_i = \mathrm{\bf v}^\top \mathrm{tanh}(\mathrm{\bf W}_1 h^A_i + \mathrm{\bf W}_2 h^A_t + \mathrm{\bf W}_3 \tilde{h}^A_i),
\end{equation}
\begin{equation}
    s^t_i = \mathrm{Softmax}(a^t_i).
\end{equation}
\end{small}
\noindent $\tilde{h}^A_i$ denotes the previous history-aware aspect representation (refer to Eq. \ref{eq:haar}). $\mathrm{\bf v} \in \mathbb{R}^{2\mathrm{dim}^A_h}$ can be learned during training. $\mathrm{\bf W}_{1,2,3} \in \mathbb{R}^{2\mathrm{dim}^A_h \times 2\mathrm{dim}^A_h}$ are parameters associated with previous aspect representations, current aspect representation and previous history-aware aspect representations respectively. 
Then, the aspect history $\hat{h}^A_t$ is obtained as follows:
\begin{small}
\begin{equation}
    \hat{h}^A_t =\sum^{t-1}_{i=t-N^A} s^t_i \times \tilde{h}^A_i.
\end{equation}
\end{small}

\noindent To benefit from the previous aspect detection, we consolidate the hidden aspect representation with the distilled aspect history to generate features for the current prediction. Specifically, we adopt a way similar to the residual block~\cite{he2016deep}, which is shown to be useful in refining word-level features in Machine Translation~\cite{wu2016google} and Part-Of-Speech tagging~\cite{bjerva2016semantic}, to calculate the history-aware aspect representations $\tilde{h}^A_t$ at the time step $t$:
\begin{equation}
\label{eq:haar}
    \tilde{h}^A_t = h^A_t + \mathrm{ReLU}(\hat{h}^A_t),
\end{equation}
where ReLU is the relu activation function. 

\subsubsection{Capturing Opinion Summary}
Previous works show that modeling aspect-opinion association is helpful to improve the accuracy of ATE, as exemplified in employing attention mechanism for calculating the opinion information \cite{wang2017coupled,li-lam:2017:EMNLP2017}. MIN \cite{li-lam:2017:EMNLP2017} focuses on a few surrounding opinion representations and computes their importance scores according to the proximity and the opinion salience derived from a given opinion lexicon. However, it is unable to capture the long-range association between aspects and opinions. Besides, the association is not strong because only the distance information is modeled.
Although CMLA \cite{wang2017coupled} can exploit global opinion information for aspect extraction, it may suffer from the noise brought in by attention-based feature aggregation. Taking the aspect term ``fish'' in \textit{``Furthermore, while the \textbf{fish} is unquestionably fresh, \textbf{rolls} tend to be inexplicably bland.''} as an example, it might be enough to tell ``fish'' is an aspect given the appearance of the strongly related opinion ``fresh''. However, CMLA employs conventional attention and does not have a mechanism to suppress the noise caused by other terms such as ``rolls''. Dependency parsing seems to be a good solution for finding the most related opinion and indeed it was utilized in \cite{wang-EtAl:2016:EMNLP20164}, but the parser is prone to generating mistakes when processing the informal online reviews, as discussed in~\cite{li-lam:2017:EMNLP2017}.

To make use of opinion information and suppress the possible noise, we propose a novel Selective Transformation Network (STN) (the \textbf{STN} block in Figure~\ref{fig:architecture}), and insert it before attending to global opinion features so that more important features with respect to a given aspect candidate will be highlighted. Specifically, STN first calculates a new opinion representation $\Hat{h}^O_{i,t}$ given the current aspect feature $\tilde{h}^A_t$ as follows:
\begin{equation}
\label{eq:new_op_re}
   \Hat{h}^O_{i,t} = h^O_i + \mathrm{ReLU}(\mathrm{\bf W}_4 \tilde{h}^A_t + \mathrm{\bf W}_5 h^O_i),
\end{equation}
where $\mathrm{\bf W}_{4}$ and $\mathrm{\bf W}_{5} \in \mathbb{R}^{2\mathrm{dim}^O_h \times 2\mathrm{dim}^O_h}$ are parameters for history-aware aspect representations and opinion representations respectively. They map $\tilde{h}^A_t$ and $h^O_i$ to the same subspace. Here the aspect feature $\tilde{h}^A_t$ acts as a ``filter'' to keep more important opinion features. Equation \ref{eq:new_op_re} also introduces a residual block to obtain a better opinion representation $\Hat{h}^O_{i,t}$, which is conditioned on the current aspect feature $\tilde{h}^A_t$. 

For distilling the global opinion summary, we introduce a bi-linear term to calculate the association score between $\tilde{h}^A_t$ and each $\Hat{h}^O_{i,t}$: 
\begin{equation}
\label{eq:w_i_t}
    w_{i,t} = \mathrm{Softmax}(\mathrm{tanh}(\tilde{h}^A_t \mathrm{\bf W}_{bi} \Hat{h}^O_{i,t} + \mathrm{\bf b}_{bi})), 
\end{equation}
where $\mathrm{\bf W}_{bi}$ and $\mathrm{\bf b}_{bi}$ are parameters of the Bi-Linear Attention layer. The improved opinion summary $\Hat{h}^O_t$ at the time $t$ is obtained via the weighted sum of the opinion representations:
\begin{equation}
    \Hat{h}^O_t = \sum^{T}_{i=1} w_{i,t} \times \Hat{h}^O_{i, t}.
\end{equation}
Finally, we concatenate the opinion summary $\Hat{h}^O_t$ and the history-aware aspect representation $\tilde{h}^A_t$ and feed it into the top-most fully-connected (FC) layer for aspect prediction:
\begin{equation}
    f^A_t = [\tilde{h}^A_t : \Hat{h}^O_t],
\end{equation}
\begin{equation}
    P(y^A_{t}|x_t) = \mathrm{Softmax}(\mathrm{\bf W}^A_{f} f^A_t + \mathrm{\bf b}^A_f).
\end{equation}
Note that our framework actually performs a multi-task learning, i.e. predicting both aspects and opinions. We regard the initial token-level representations $h^O_i$ as the features for opinion prediction: 
\begin{equation}
    P(y^O_i|x_i) = \mathrm{Softmax}(\mathrm{\bf W}^O_{f} h^O_i + \mathrm{\bf b}^O_f).
\end{equation}
$\mathrm{\bf W}^{\mathcal{T}}_{f}$ and $\mathrm{\bf b}^{\mathcal{T}}_f$ are parameters of the FC layers.



\subsection{Joint Training}
All the components in the proposed framework are differentiable. Thus, our framework can be efficiently trained with gradient methods. We use the token-level cross-entropy error between the predicted distribution $P(y^{\mathcal{T}}_t|x_t)$ ($\mathcal{T} \in \{A, O\}$) and the gold distribution $P(y^{\mathcal{T}, g}_t|x_t)$ as the loss function:
\begin{equation}\
\begin{split}
\mathcal{L}_{\mathcal{T}} &= -\frac{1}{~T~}\sum^{T}_{t=1} P(y^{\mathcal{T}, g}_{t}|x_{t}) \odot \log [P(y^{\mathcal{T}}_{t}|x_{t})]. 
\end{split}
\end{equation}
Then, the losses from both tasks are combined to form the training objective of the entire model:
\begin{equation}
\mathcal{J}(\theta)=\mathcal{L}_A+\mathcal{L}_O,
\end{equation}
where $\mathcal{L}_A$ and $\mathcal{L}_O$ represent the loss functions for aspect and opinion extractions respectively.

\section{Experiment}
\subsection{Datasets}
To evaluate the effectiveness of the proposed framework for the ATE task, we conduct experiments over four benchmark datasets from the SemEval ABSA challenge~\cite{pontiki-EtAl:2014:SemEval,pontiki-EtAl:2015:SemEval,pontiki-EtAl:2016:SemEval}. Table~\ref{dataset_statistic} shows their statistics. $D_1$ (SemEval 2014) contains reviews of the laptop domain and those of $D_2$ (SemEval 2014), $D_3$ (SemEval 2015) and $D_4$ (SemEval 2016) are for the restaurant domain. In these datasets, aspect terms have been labeled by the task organizer.

\begin{table}[ht!]
\centering
\resizebox{0.43\textwidth}{!}{%
\begin{tabular}{c|c|c|c|c}
\Xhline{3\arrayrulewidth}
 \multicolumn{2}{c|}{} & \# Sentences & \# Aspects & \# Sentences with aspects \\ \hline
\multirow{2}{*}{$D_1$} & TRAIN & 3045 & 2358 & 1484 \\ \cline{2-5}
& TEST & 800 & 654 & 422 \\ \hline
\multirow{2}{*}{$D_2$} & TRAIN & 3041 & 1743 & 1020\\ \cline{2-5}
& TEST & 800 & 1134 & 194 \\ \hline
\multirow{2}{*}{$D_3$} & TRAIN & 1315 & 1192 & 832\\ \cline{2-5}
& TEST & 685 & 542 & 401\\ \hline
\multirow{2}{*}{$D_4$} & TRAIN & 2000 & 1743 & 1233\\ \cline{2-5}
& TEST & 676 & 622 & 420 \\ \Xhline{3\arrayrulewidth}
\end{tabular}}
\caption{Statistics of datasets.}
\label{dataset_statistic}
\end{table}

Gold standard annotations for opinion words are not provided. Thus, we choose words with strong subjectivity from MPQA\footnote{http://mpqa.cs.pitt.edu/} to provide the distant supervision~\cite{Mintz_distantIE}. To compare with the best SemEval systems and the current state-of-the-art methods, we use the standard train-test split in SemEval challenge as shown in Table~\ref{dataset_statistic}.

\subsection{Comparisons}
We compare our framework with the following methods:
\begin{itemize}
\item \textbf{CRF-1}: Conditional Random Fields with basic feature templates\footnote{http://sklearn-crfsuite.readthedocs.io/en/latest/}.
\item \textbf{CRF-2}: Conditional Random Fields with basic feature templates and word embeddings.
\item \textbf{Semi-CRF}: First-order Semi-Markov Conditional Random Fields~\cite{sarawagi2004semi} and the feature templates in \citet{cuong2014conditional} are adopted. 
\item \textbf{LSTM}: Vanilla bi-directional LSTM with pre-trained word embeddings.
\item \textbf{IHS\_RD}~\cite{chernyshevich:2014:SemEval}, \textbf{DLIREC}~\cite{toh-wang:2014:SemEval},  \textbf{EliXa}~\cite{sanvicente-saralegi-agerri:2015:SemEval}, \textbf{NLANGP}~\cite{toh-su:2016:SemEval}: The winning systems in the ATE subtask in SemEval ABSA challenge~\cite{pontiki-EtAl:2014:SemEval,pontiki-EtAl:2015:SemEval,pontiki-EtAl:2016:SemEval}.
\item \textbf{WDEmb}~\cite{yin2016unsupervised}: Enhanced CRF with word embeddings, dependency path embeddings and linear context embeddings.
\item \textbf{MIN}~\cite{li-lam:2017:EMNLP2017}: MIN consists of three LSTMs. Two LSTMs are employed to model the memory interactions between ATE and opinion detection. The last one is a vanilla LSTM used to predict the subjectivity of the sentence as additional guidance.  
\item \textbf{RNCRF}~\cite{wang-EtAl:2016:EMNLP20164}: CRF with high-level representations learned from Dependency Tree based Recursive Neural Network.
\item \textbf{CMLA}~\cite{wang2017coupled}: CMLA is a multi-layer architecture where each layer consists of two coupled GRUs to model the relation between aspect terms and opinion words.
\end{itemize}

To clarify, our framework aims at extracting aspect terms where the opinion information is employed as auxiliary, while RNCRF and CMLA perform joint extraction of aspects and opinions. Nevertheless, the comparison between our framework and RNCRF/CMLA is still fair, because we do not use manually annotated opinions as used by RNCRF and CMLA, instead, we employ an existing opinion lexicon to provide weak opinion supervision.

\subsection{Settings}
We pre-processed each dataset by lowercasing all words and replace all punctuations with \texttt{PUNCT}. We use pre-trained GloVe 840B vectors\footnote{https://nlp.stanford.edu/projects/glove/}~\cite{pennington2014glove} to initialize the word embeddings and the dimension (i.e., $\mathrm{dim}_w$) is 300. For out-of-vocabulary words, we randomly sample their embeddings from the uniform distribution $\mathcal{U}(-0.25, 0.25)$ as done in~\cite{kim:2014:EMNLP2014}. All of the weight matrices except those in LSTMs are initialized from the uniform distribution $\mathcal{U}(-0.2, 0.2)$. For the initialization of the matrices in LSTMs, we adopt Glorot Uniform strategy~\cite{glorot2010understanding}. Besides, all biases are initialized as 0's. 

The model is trained with SGD. We apply dropout over the ultimate aspect/opinion features and the input word embeddings of LSTMs. The dropout rates are empirically set as 0.5.
With 5-fold cross-validation on the training data of $D_2$, other hyper-parameters are set as follows: $dim^A_h=100$, $dim^O_h=30$; the number of cached historical aspect representations $N^A$ is 5; the learning rate of SGD is 0.07.


\begin{table}[!t]
    \centering
    \resizebox{0.48\textwidth}{!}{%
    \begin{tabular}{lcccc}
    \Xhline{3\arrayrulewidth}
         Models & $D_1$ & $D_2$ & $D_3$ & $D_4$ \\ \hline \hline
         CRF-1 & 72.77 & 79.72 & 62.67 & 66.96 \\ 
         CRF-2 & 74.01 & 82.33 & 67.54 & 69.56 \\ 
         Semi-CRF & 68.75 & 79.60 & 62.69 & 66.35 \\ 
         LSTM & 75.71 & 82.01 & 68.26 & 70.35 \\ 
         IHS\_RD (\textbf{$D_1$ winner}) & 74.55 & 79.62 & - & - \\ 
         DLIREC (\textbf{$D_2$ winner}) & 73.78 & 84.01 & - & - \\ 
         EliXa (\textbf{$D_3$ winner}) & - & - & 70.04 & - \\ 
         NLANGP (\textbf{$D_4$ winner}) & - & - & 67.12 & 72.34 \\ 
         WDEmb & 75.16 & 84.97 & 69.73 & - \\ 
         MIN & 77.58 & - & - & 73.44 \\ 
         RNCRF & 78.42 & 84.93 & 67.74$^{\natural}$ & 69.72* \\
         CMLA & 77.80 & 85.29 & 70.73 & 72.77* \\ \hline \hline
         OURS w/o \textbf{THA} & 77.64 & 84.30 & 70.89 & 72.62 \\ 
         OURS w/o \textbf{STN} & 77.45 & 83.88 & 70.09 & 72.18 \\ 
         OURS w/o \textbf{THA} \& \textbf{STN} & 76.95 & 83.48 & 69.77 & 71.87 \\ \hline \hline
         OURS & \textbf{79.52} & \textbf{85.61} & \textbf{71.46} & \textbf{73.61} \\ \Xhline{3\arrayrulewidth}
    \end{tabular}}
    \caption{Experimental results ($F_1$ score, \%). The first four methods are implemented by us, and other results without markers are copied from their papers. The results with `*' are reproduced by us with the released code by the authors. For RNCRF, the result with `$^{\natural}$' is copied from the paper of CMLA (they have the same authors). `-' indicates the results were not available in their papers. }
    \label{tab:main_results}
\end{table}

\subsection{Main Results}
As shown in Table~\ref{tab:main_results}, the proposed framework consistently obtains the best scores on all of the four datasets. Compared with the  winning systems of SemEval ABSA, our framework achieves 5.0\%, 1.6\%, 1.4\%, 1.3\% absolute gains on $D_1$, $D_2$, $D_3$ and $D_4$ respectively.

Our framework can outperform RNCRF, a state-of-the-art model based on dependency parsing, on all datasets. We also notice that RNCRF does not perform well on $D_3$ and $D_4$ (3.7\% and 3.9\% inferior than ours). We find that $D_3$ and $D_4$ contain many informal reviews, thus RNCRF's performance degradation is probably due to the errors from the dependency parser when processing such informal texts.

CMLA and MIN do not rely on dependency parsing, instead, they employ attention mechanism to distill opinion information to help aspect extraction. Our framework consistently performs better than them. The gains presumably come from two perspectives: (1) In our model, the opinion summary is exploited after performing the selective transformation conditioned on the current aspect features, thus the summary can to some extent avoid the noise due to directly applying conventional attention. (2) Our model can discover some uncommon aspects under the guidance of some commonly-used aspects in coordinate structures by the history attention.

CRF with basic feature template is not strong, therefore, we add CRF-2 as another baseline. As shown in Table~\ref{tab:main_results}, CRF-2 with word embeddings achieves much better results than CRF-1 on all datasets. WDEmb, which is also an enhanced CRF-based method using additional dependency context embeddings, obtains superior performances than \mbox{CRF-2}. Therefore, the above comparison shows that word embeddings are useful and the embeddings incorporating structure information can further improve the performance.

\begin{figure*}
    \centering
    \begin{minipage}{0.88\textwidth}
    \begin{minipage}{0.48\textwidth}
        \centering
        \includegraphics[height=27mm,width=1\textwidth]{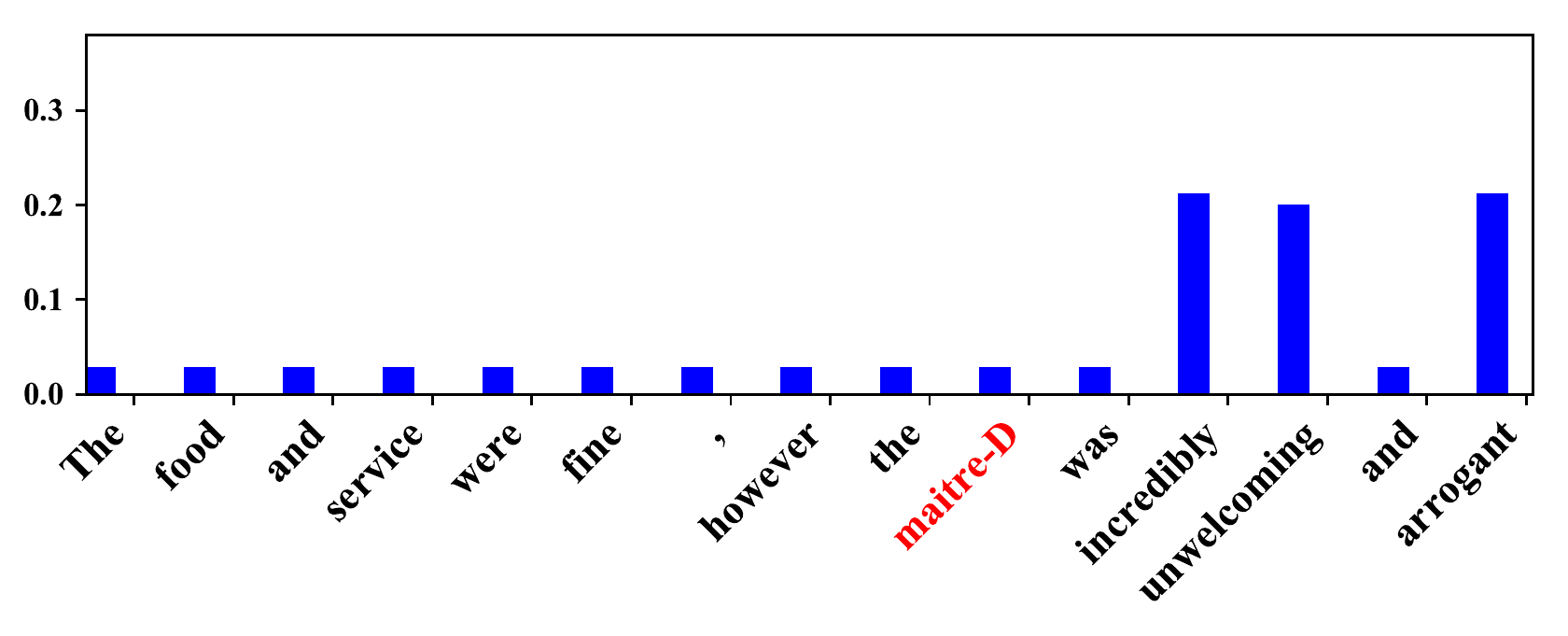}
        \subcaption{Scores generated by our framework}
    \label{fig:1a}
    \end{minipage}%
   \begin{minipage}{0.48\textwidth}
      \centering
        \includegraphics[height=27mm,width=1\textwidth]{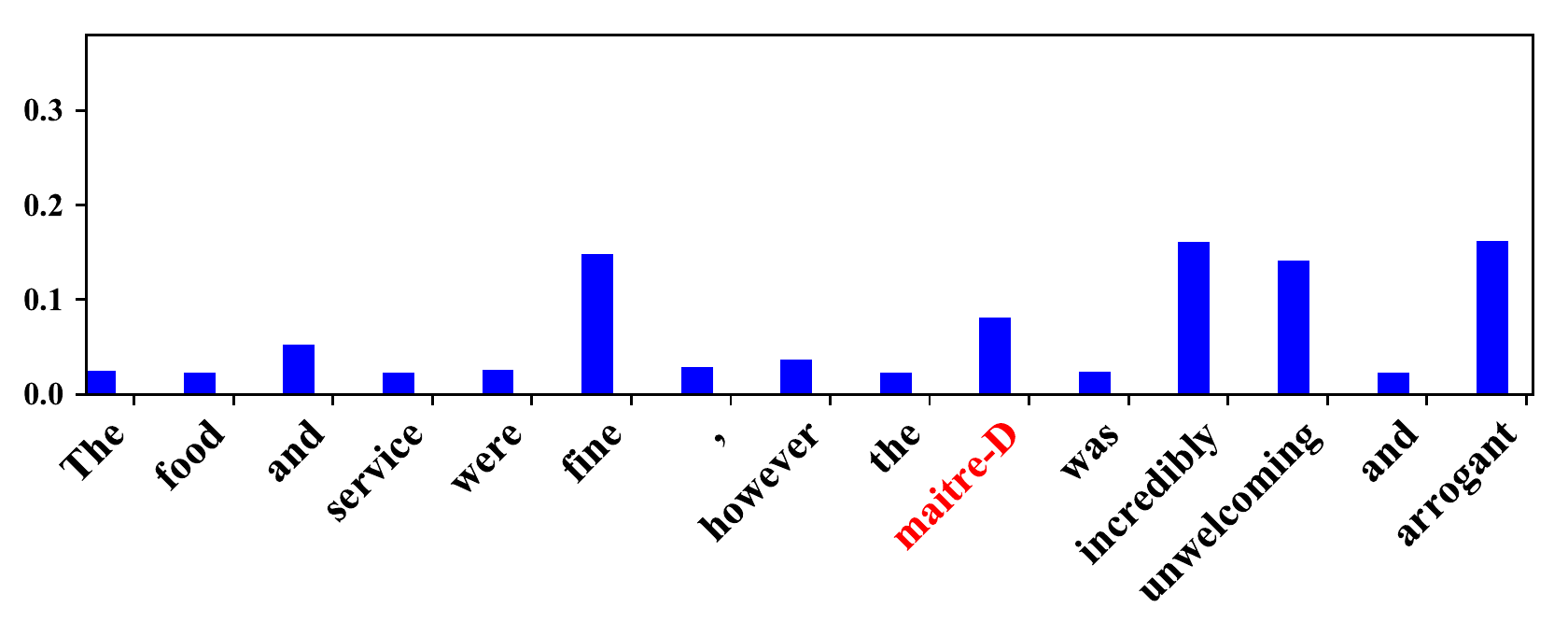}
        \subcaption{Scores generated by ``OURS w/o STN''.}
        \label{fig:1b}
    \end{minipage}%
    
    \end{minipage}
    \begin{minipage}{0.88\textwidth}
    \begin{minipage}{0.48\textwidth}
        \centering
        \includegraphics[height=25mm,width=1\textwidth]{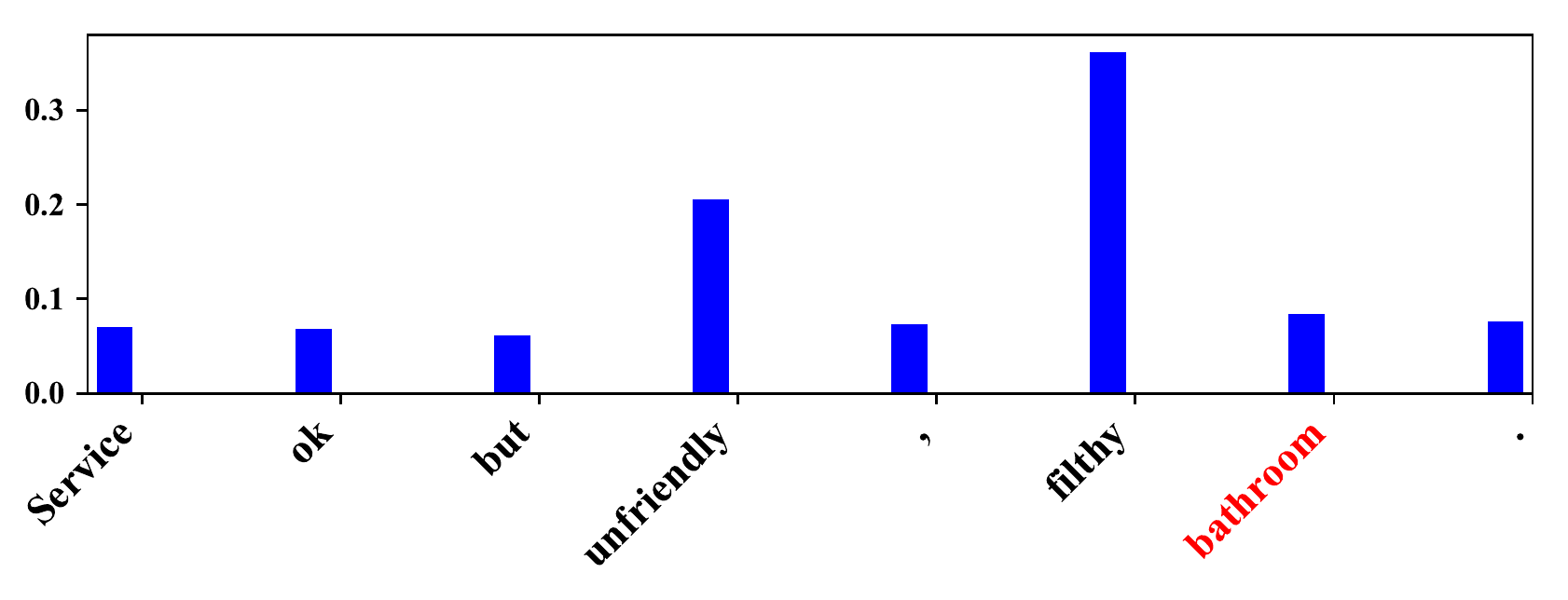}
        \subcaption{Scores generated by our framework}
    \label{fig:1c}
    \end{minipage}
   \begin{minipage}{0.48\textwidth}
  \centering
\includegraphics[height=25mm,width=1\textwidth]{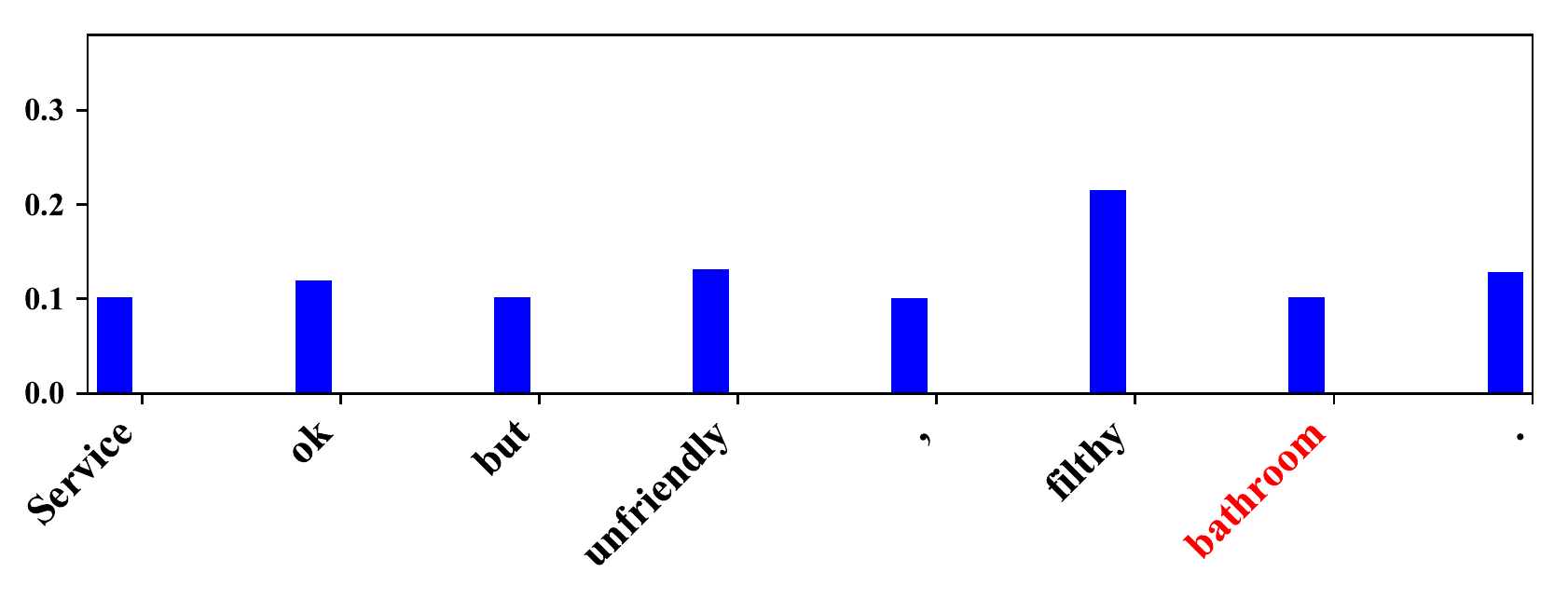}
\subcaption{Scores generated by ``OURS w/o STN''.}
\label{fig:1d}
\end{minipage}
    \caption{Opinion attention scores (i.e. $w_{i,t}$ in Equation \ref{eq:w_i_t}) with respect to ``maitre-D'' and ``bathroom''.}
    \label{fig:my_label}
    \end{minipage}
\end{figure*}

\subsection{Ablation Study}
To further investigate the efficacy of the key components in our framework, namely, \textbf{THA} and \textbf{STN}, we perform ablation study as shown in the second block of Table~\ref{tab:main_results}. 
The results show that each of THA and STN is helpful for improving the performance, and the contribution of STN is slightly larger than THA. 
``OURS w/o THA \& STN'' only keeps the basic bi-linear attention. Although it performs not bad, it is still less competitive compared with the strongest baseline (i.e., CMLA), suggesting that only using attention mechanism to distill opinion summary is not enough. After inserting the STN component before the bi-linear attention, i.e. ``OURS w/o THA'', we get about 1\% absolute gains on each dataset, and then the performance is comparable to CMLA. By adding THA, i.e. ``OURS'', the performance  is further improved, and all state-of-the-art methods are surpassed.

\begin{table*}[!ht]
    \centering
    \resizebox{1\textwidth}{!}
    {
    \begin{tabular}{L{7cm}||L{4cm}|L{6cm}|L{5cm}}
    \Xhline{3\arrayrulewidth}
    \hspace{2.2cm}Input sentences & Output of LSTM & Output of OURS w/o THA \& STN & Output of OURS \\ \hline
    1. \textit{the device speaks about it self} & \textit{device} & \textbf{NONE} & \textbf{NONE} \\ \hline 
    2. \textit{Great \textcolor{red}{\underline{survice}} !} & \textbf{NONE} & \textit{survice} & \textit{survice} \\ \hline
    3. \textit{Apple is unmatched in \textcolor{red}{\underline{product quality}}, \textcolor{red}{\underline{aesthetics}}, \textcolor{red}{\underline{craftmanship}}, and \textcolor{red}{\underline{custormer service}}} & \textit{quality, aesthetics, custormer service} & \textit{quality, customer service} & \textit{product quality, aesthetics, craftmanship, custormer service} \\ \hline
    4. \textit{I am pleased with the fast \textcolor{red}{\underline{log on}}, speedy \textcolor{red}{\underline{WiFi connection}} and the long \textcolor{red}{\underline{battery life}}} & \textit{WiFi connection, battery life} & \textit{log, WiFi connection, battery life} & \textit{log on, WiFi connection, battery life} \\ \hline
     5. \textit{Also, I personally wasn't a fan of the \textcolor{red}{\underline{portobello and asparagus mole}}} & \textit{asparagus mole} & \textit{asparagus mole} & \textit{portobello and asparagus mole} \\
    \Xhline{3\arrayrulewidth}
    \end{tabular}}
    \caption{Case analysis. In the input sentences, the gold standard aspect terms are underlined and in red. }
    \label{tab:predictions}
\end{table*}

\subsection{Attention Visualization and Case Study}
In Figure \ref{fig:my_label}, we visualize the opinion attention scores of the words in two example sentences with the candidate aspects ``maitre-D'' and ``bathroom''. The scores in Figures \ref{fig:1a} and \ref{fig:1c} show that our full model captures the related opinion words very accurately with significantly larger scores, i.e. ``incredibly'', ``unwelcoming'' and ``arrogant'' for ``maitre-D'', and ``unfriendly'' and ``filthy'' for ``bathroom''. 
``OURS w/o STN'' directly applies attention over the opinion hidden states $h_i^O$'s, similar to what CMLA does. As shown in Figure \ref{fig:1b}, it captures some unrelated opinion words (e.g. ``fine'') and even some non-opinionated words. As a result, it brings in some noise into the global opinion summary, and consequently the final prediction accuracy will be affected. This example demonstrates that the proposed STN works pretty well to help attend to more related opinion words given a particular aspect.

Some predictions of our model and those of LSTM and OURS w/o THA \& STN are given in Table~\ref{tab:predictions}. The models incorporating attention-based opinion summary (i.e., OURS and OURS w/o THA \& STN) can better determine if the commonly-used nouns are aspect terms or not (e.g. ``device'' in the first input), since they make decisions based on the global opinion information. Besides, they are able to extract some infrequent or even misspelled aspect terms (e.g. ``survice'' in the second input) based on the indicative clues provided by opinion words. For the last three cases, having aspects in coordinate structures (i.e. the third and the fourth) or long aspects (i.e. the fifth), our model can give precise predictions owing to the previous detection clues captured by THA. Without using these clues, the baseline models fail. 

\section{Related Work}
Some initial works \cite{hu2004miningA} developed a bootstrapping framework for tackling Aspect Term Extraction (ATE) based on the observation that opinion words are usually located around the aspects. \cite{popescu-etzioni:2005:HLTEMNLP} and \cite{qiu2011opinion} performed co-extraction of aspect terms and opinion words based on sophisticated syntactic patterns. However, relying on syntactic patterns suffers from parsing errors when processing informal online reviews.
To avoid this drawback, \cite{liu-xu-zhao:2012:EMNLP-CoNLL,liu2013opinion} employed word-based translation models. Specifically, these models formulated the ATE task as a monolingual word alignment process and aspect-opinion relation is captured by alignment links rather than word dependencies.
The ATE task can also be formulated as a token-level sequence labeling problem. The winning systems~\cite{chernyshevich:2014:SemEval,sanvicente-saralegi-agerri:2015:SemEval,toh-su:2016:SemEval} of SemEval ABSA challenges employed traditional sequence models, such as Conditional Random Fields (CRFs) and Maximum Entropy (ME), to detect aspects. Besides heavy feature engineering, they also ignored the consideration of opinions. 


Recently, neural network based models, such as LSTM-based \cite{liu-joty-meng:2015:EMNLP} and CNN-based \cite{poria2016aspect} methods, become the mainstream approach. Later on, some neural models jointly extracting aspect and opinion were proposed. \cite{wang-EtAl:2016:EMNLP20164} performs the two task in a single Tree-Based Recursive Neural Network. Their network structure depends on dependency parsing, which is prone to error on informal reviews. CMLA~\cite{wang2017coupled} consists of multiple attention layers on top of standard GRUs to extract the aspects and opinion words. Similarly, MIN~\cite{li-lam:2017:EMNLP2017} employs multiple LSTMs to interactively perform aspect term extraction and opinion word extraction in a multi-task learning framework. Our framework is different from them in two perspectives: (1) It filters the opinion summary by incorporating the aspect features at each time step into the original opinion representations; 
(2) It exploits history information of aspect detection to capture the coordinate structures and previous aspect features.

\section{Concluding Discussions}
For more accurate aspect term extraction, we explored two important types of information, namely aspect detection history, and opinion summary. We design two components, i.e. truncated history attention, and selective transformation network. 
Experimental results show that our model dominates those joint extraction works such as RNCRF and CMLA on the performance of ATE. It suggests that the joint extraction sacrifices the accuracy of aspect prediction, although the ground-truth opinion words were annotated by these authors. Moreover, one should notice that those joint extraction methods do not care about the correspondence between the extracted aspect terms and opinion words. Therefore, the necessity of such joint extraction should be obelized, given the experimental findings in this paper. 


\begin{small}
\bibliographystyle{named}
\bibliography{ijcai18}
\end{small}

\end{document}